\documentclass[conference]{IEEEconf}
\IEEEoverridecommandlockouts    % to override locked commands

\usepackage{multirow}
\usepackage{lipsum}
\usepackage{amsmath}
\usepackage{amssymb}
\usepackage[ruled,vlined]{algorithm2e}
\usepackage{graphicx}
\usepackage{siunitx}
\usepackage{cite}
\usepackage{bm}
\usepackage[acronym]{glossaries}
\usepackage{url}       
\usepackage{hyperref}
\usepackage{censor}
\usepackage{booktabs}
\usepackage{anyfontsize}
\usepackage{placeins}
\usepackage{tikz}

\newacronym{mpc}{MPC}{Model Predictive Control}
\newacronym{sgpr}{SGPR}{Sparse Gaussian Process Regression}
\newacronym{gpr}{GPR}{Gaussian Process Regression}
\newacronym{blr}{BLR}{Bayesian Linear Regression}
\newacronym{rbf}{RBF}{Radial Basis Function}
\newacronym{pid}{PID}{Proportional-Integral-Derivative}
\newacronym{sqp}{SQP}{Sequential Quadratic Programming}
\newcommand{\vect}[1]{\boldsymbol{#1}}

\newcommand{\dt}{\Delta t}
\newglossaryentry{dt}{
    name={$\dt$},
    description={Sampling time in seconds}
}

\newcommand{\steer}{\delta}
\newglossaryentry{steer}{
    name={$\vect{\steer}$},
    description={Steering angle sequence}
}

\newcommand{\velocity}{v}

\newcommand{\orientation}{\varphi}
\newglossaryentry{orientation}{
    name={$\orientation$},
    description={Vehicle orientation}
}

\newcommand{\state}{\mu}
\newglossaryentry{state}{
    name={$\vect{\state}_k$},
    description={Vehicle state vector on step $k$}
}

\newcommand{\predictionIndex}{k}
\newglossaryentry{predictionIndex}{
    name={$\predictionIndex$},
    description={Prediction index}
}

\newcommand{\predictionCovariance}[1]{\IfNoValueTF{#1}{\vect{\Sigma}}{\vect{\Sigma}_{#1}}}
\newglossaryentry{predictionCovariance}{
    name={$\vect{\Sigma}_{\predictionIndex}$},
    description={Prediciton covariance on prediction step $\predictionIndex$}
}

\newcommand{\timeIndex}{t}
\newglossaryentry{timeIndex}{
    name={$\timeIndex$},
    description={Time index}
}

\newcommand{\iz}{I_z}
\newglossaryentry{iz}{
    name={$\iz$},
    description={Moment of inertia of the vehicle on the z axis}
}

\newcommand{\fx}{F_x}
\newglossaryentry{fx}{
    name={$\fx$},
    description={Accelerating force}
}

\newcommand{\fxmax}{F_{x, \max}}
\newglossaryentry{fxmax}{
    name={$\fxmax$},
    description={Maximum accelerating force}
}

\newcommand{\mass}{m}
\newglossaryentry{mass}{
    name={$\mass$},
    description={Vehicle mass}
}

\newcommand{\lF}{l_F}
\newglossaryentry{lF}{
    name={$\lF$},
    description={Distance between center of gravity and the front axle}
}

\newcommand{\lR}{l_R}
\newglossaryentry{lR}{
    name={$\lR$},
    description={Distance between center of gravity and the rear axle}
}

\newcommand{\predictionStepAmmount}{K}
\newglossaryentry{predictionStepAmmount}{
    name={$\predictionStepAmmount$},
    description={Number of prediction steps}
}

\newcommand{\progress}[1]{\theta_{#1}}
\newglossaryentry{progress}{
    name={$\progress{\predictionIndex}$},
    description={Progress variable (Nearest point on the reference path at prediction $\predictionIndex$)}
}

\newcommand{\rx}[1]{R_x({#1})}
\newglossaryentry{rx}{
    name={$\rx{\progress{\predictionIndex}}$},
    description={x-position of the reference path on progress variable $\progress{\predictionIndex}$}
}

\newcommand{\ry}[1]{R_y({#1})}
\newglossaryentry{ry}{
    name={$\ry{\progress{\predictionIndex}}$},
    description={y-position of the reference path on progress variable $\progress{\predictionIndex}$}
}

\newcommand{\ro}{R_\varphi(\progress{\predictionIndex})}
\newglossaryentry{ro}{
    name={$\ro$},
    description={Orientation of the reference path on progress variable $\progress{\predictionIndex}$}
}

\newcommand{\rw}{R_w(\progress{\predictionIndex})}
\newglossaryentry{rw}{
    name={$\rw$},
    description={Width of the reference path on progress variable $\progress{\predictionIndex}$}
}

\newcommand{\targetSpeed}[1]{R_v(\theta_{#1})}
\newglossaryentry{targetSpeed}{
    name={$\targetSpeed{\predictionIndex}$},
    description={Target speed on the reference path on progress variable $\theta_\predictionIndex$}
}

\newcommand{\mappingMatrix}{B}
\newglossaryentry{mappingMatrix}{
    name={$\vect{\mappingMatrix}$},
    description={Veloctiy mapping matrix}
}

\newcommand{\ppTube}{\psi_{\textnormal{tube}}}
\newglossaryentry{ppTube}{
    name={$\ppTube$},
    description={Pure pursuit tube size}
}

\newcommand{\ppDistance}{\psi_{\textnormal{dist}}}
\newglossaryentry{ppDistance}{
    name={$\ppDistance$},
    description={Pure pursuit target point distance}
}

\newcommand{\datapoint}{z}
\newglossaryentry{datapoint}{
    name={$\vect{\datapoint}$},
    description={Correction model datapoint}
}

\newcommand{\contouringError}{e_{c, k}}
\newglossaryentry{contouringError}{
    name={$\contouringError$},
    description={Contouring error on prediction $\predictionIndex$}
}

\newcommand{\lagError}{e_{l, k}}
\newglossaryentry{lagError}{
    name={$\lagError$},
    description={Lag error on prediction $\predictionIndex$}
}

\newcommand{\contouringErrorWeight}{\eta_c}
\newglossaryentry{contouringErrorWeight}{
    name={$\contouringErrorWeight$},
    description={Contouring error weight}
}

\newcommand{\lagErrorWeight}{\eta_l}
\newglossaryentry{lagErrorWeight}{
    name={$\lagErrorWeight$},
    description={Lag error weight}
}

\newcommand{\steeringErrorWeight}{\eta_\delta}
\newglossaryentry{steeringErrorWeight}{
    name={$\steeringErrorWeight$},
    description={Steering error weight}
}

\newcommand{\stabilityErrorWeight}{\eta_{\Delta v_r}}
\newglossaryentry{stabilityErrorWeight}{
    name={$\stabilityErrorWeight$},
    description={Stability error weight}
}

\newcommand{\borderErrorWeight}{\eta_b}
\newglossaryentry{borderErrorWeight}{
    name={$\borderErrorWeight$},
    description={Border error weight}
}

\newcommand{\onlineModelInsertionThreshold}{\xi_\textnormal{insertion}}
\newglossaryentry{onlineModelInsertionThreshold}{
    name={$\onlineModelInsertionThreshold$},
    description={Online model insertion threshold}
}

\newcommand{\onlineModelOutlierThreshold}{\xi_\textnormal{outlier}}
\newglossaryentry{onlineModelOutlierThreshold}{
    name={$\onlineModelOutlierThreshold$},
    description={Online model outlier threshold}
}

\newcommand{\onlineModelQuantile}{\xi_\textnormal{quantile}}
\newglossaryentry{onlineModelQuantile}{
    name={$\onlineModelQuantile$},
    description={Online model gaussian process probability quantile}
}

\newcommand{\velocityControlParameter}{\tau_p}
\newglossaryentry{velocityControlParameter}{
    name={$\velocityControlParameter$},
    description={Velocity controller P-component}
}

\newcommand{\throttle}{u}
\newglossaryentry{throttle}{
    name={$\throttle_k$},
    description={Throttle command at prediction step $k$}
}

\newcommand{\steerMax}{\steer_{\max}}
\newglossaryentry{steerMax}{
    name={$\steerMax$},
    description={Maximum steering angle}
}

\newcommand{\steerMin}{\steer_{\min}}
\newglossaryentry{steerMin}{
    name={$\steerMin$},
    description={Minimum steering angle}
}

\newcommand{\trustRegionSize}{\phi}
\newglossaryentry{trustRegionSize}{
    name={$\trustRegionSize$},
    description={Trust region size}
}

\newcommand{\iterationIndex}{j}
\newglossaryentry{iterationIndex}{
    name={$\iterationIndex$},
    description={Index of the current optimization iteration}
}

\newcommand{\correctionVariable}{\gamma}
\newglossaryentry{correctionVariable}{
    name={$\correctionVariable$},
    description={Correction variable for better discretization}
}

\newglossaryentry{dictionaryPoints}{
    name={$\vect{Z}$},
    description={Set of all data points in the dictionary of the online model}
}

\newcommand{\borderConfidence}{p}
\newglossaryentry{borderConfidence}{
    name={$\borderConfidence$},
    description={Confidence threshold of the border error}
}

\newcommand{\deltavx}{\dot{\velocity}_x}
\newglossaryentry{deltavx}{
    name={$\deltavx$},
    description={Current changing rate of the longitudinal velocity}
}

\newcommand{\deltasteer}{\dot{\steer}}
\newglossaryentry{deltasteer}{
    name={$\deltasteer$},
    description={Current changing rate of the steering angle}
}

\newcommand{\trustRegionIncreaseFactor}{a_\textnormal{i}}
\newglossaryentry{trustRegionIncreaseFactor}{
    name={$\trustRegionIncreaseFactor$},
    description={Increase factor of the trust region}
}

\newcommand{\trustRegionDecreaseFactor}{a_\textnormal{d}}
\newglossaryentry{trustRegionDecreaseFactor}{
    name={$\trustRegionDecreaseFactor$},
    description={Decrease factor of the trust region}
}

\newcommand{\dictionaryImportanceScore}{\gamma_i}
\newglossaryentry{dictionaryImportanceScore}{
    name={$\dictionaryImportanceScore$},
    description={Score indicating how important a data point $i$ is for the Gaussian Process}
}

\newcommand{\gprTimeWeighting}{h}
\newglossaryentry{gprTimeWeighting}{
    name={$\gprTimeWeighting$},
    description={Age weight of a data point for dictionary point replacement}
}

\newcommand{\offlineModelDifference}{\varepsilon}
\newglossaryentry{offlineModelDifference}{
    name={$\vect{\offlineModelDifference}$},
    description={Difference between true velocities and nominal model predictions}
}

\newcommand{\onlineModelDifference}{\hat{\varepsilon}}
\newglossaryentry{onlineModelDifference}{
    name={$\vect{\onlineModelDifference}$},
    description={Difference between true velocities and the offline model corrected predictions}
}

\newcommand\copyrighttext{%
  \footnotesize \textcopyright 2026 IEEE. Personal use of this material is permitted.
  Permission from IEEE must be obtained for all other uses, in any current or future
  media, including reprinting/republishing this material for advertising or promotional
  purposes, creating new collective works, for resale or redistribution to servers or
  lists, or reuse of any copyrighted component of this work in other works.
  DOI: \href{https://ieeexplore.ieee.org/document/11624038}{10.1109/IV66570.2026.11624038}}
\newcommand\copyrightnotice{%
\begin{tikzpicture}[remember picture,overlay]
\node[anchor=south,yshift=10pt] at (current page.south)
  {\fbox{\parbox{\dimexpr\textwidth-\fboxsep-\fboxrule\relax}{\copyrighttext}}};
\end{tikzpicture}%
}

\title{\LARGE \bf Vehicle Prediction Model for Enhanced MPC Path Tracking\\ in Formula Student Driverless}

 \author{
 	\parbox{\textwidth}{%
 		\centering
 		Sebastian Baader, Tamara Bergerhoff, Pascal Meißner, Frank Deinzer % Sebastian Baader$^{1}$
 	}%
 	\thanks{$^{*}$This work was supported by Mainfranken Racing e.V.}
 	\thanks{All authors are with the Center for Artificial Intelligence and Robotics (CAIRO); TUAS Würzburg-Schweinfurt, Germany}
	\thanks{Corresponding author: {\tt\small sebastian.baader@thws.de}}%
 }

\begin{document}
	
	\maketitle
	\thispagestyle{empty}
	\pagestyle{empty}
	%
	
	%%%%%%%%%%%%%%%%%%%%%%%%%%%%%%%%%%%%%%%%%%%%%%%%%%%%%%%%%%%%%%%%%%
	\begin{abstract}
		% Write one statement about the general problem and the "why" (Why) [glaube warum ist das thema interessant/wichtig]
Autonomous race cars, such as in Formula Student Driverless, operate close to their physical handling limits. The resulting highly nonlinear vehicle behavior increases the path tracking complexity, especially on narrow tracks.
%
% Tell what this paper is about (Which problem)
%Model predictive control is commonly used to address this issue, but its performance depends heavily on the accuracy of the underlying prediction model. 
Model Predictive Control (MPC) is commonly used to address this issue, a method whose performance is closely tied to the accuracy of the underlying prediction model. 
This paper presents a novel, real-time capable prediction model for autonomous race cars that adjusts to changing conditions by combining information from past runs and the current driving situation.
%
% Describe how it solves the problem (How is it addressed)
Our model is divided into three consecutive submodels: a nominal Kinematic Bicycle Model, an offline \gls{blr} model, and an online \gls{sgpr} model.
%
% Emphasize what is new or better
The proposed approach enables efficient integration of all available data without significantly increasing computational cost, ensuring high prediction accuracy and a quantitative uncertainty assessment right from the start of the run.
%
% Mention the evidence indicating the advantages of the proposed approach (What is achieved) [TODO den teil am ende wahrscheinlich nochmal überarbeiten wenn evaluierung besteht]
Compared to existing approaches, an improvement in prediction accuracy of up to \SI{57}{\percent} was achieved. Further, we successfully demonstrated the practical applicability of the model within an MPC-based path tracking controller on a real Formula Student race car.

% - Theoretisch noch vergleich bei online + offline modelle
% - Theoretisch kann mans auch ablation study nennen
% - TODO ist noch zu lang
% - TODO evtl noch auf Laufzeit Anforderungen eingehen
% - TODO evtl noch auf sicherheitsanforderungen eingehen 
% - TODO teiweise schlechte argumentation. Lieber sagen keine gute Vorhersagegenauigkeit ab begin an und keine aussage über Varianz als offline daten nicht genutzt (lieber das mit offline daten als Lösung verkaufen für das problem)
	\end{abstract}
	%%%%%%%%%%%%%%%%%%%%%%%%%%%%%%%%%%%%%%%%%%%%%%%%%%%%%%%%%%%%%%%%%%
	
	\copyrightnotice

	\section{Introduction}
	\glsresetall
%Start with a motivation (Why)

Millions of people manage to precisely follow a path with their car every day. For autonomous systems, however, this task is still a considerable challenge. In competitions such as Formula Student \cite{fsg_rules}, where vehicles operate at high speeds along narrow and winding tracks, this so-called path tracking problem is especially difficult.
%
%Formula Student is an international design competition in which teams of students develop and manufacture a fully functional race car in one year that is, among other features, equipped with autonomous driving functions (see Fig. \ref{fig:mf17}). Every year, university teams from around the world compete against each other in various disciplines, where both the technical design and the performance on the race track are evaluated. The objective is to drive as quickly as possible through a track marked by differently colored cones. These tracks are typically \SI{3}{\meter} wide, and speeds of up to \SI{100}{\kilo\metre\per\hour} can be reached.
%
High speeds lead to highly nonlinear and unpredictable driving behavior. This occurs due to significant forces acting on the vehicle. The tires operate with high slip, and the available time to react to disturbances is minimal. At such conditions there is little room for deviation from the racing line, and it requires extremely precise path tracking, because leaving the track results in penalties and bears the risk of colliding with surrounding objects like walls.

\begin{figure}[htbp]
  \vspace{0.2 cm}
  \centerline{\includegraphics[width=0.5\textwidth]{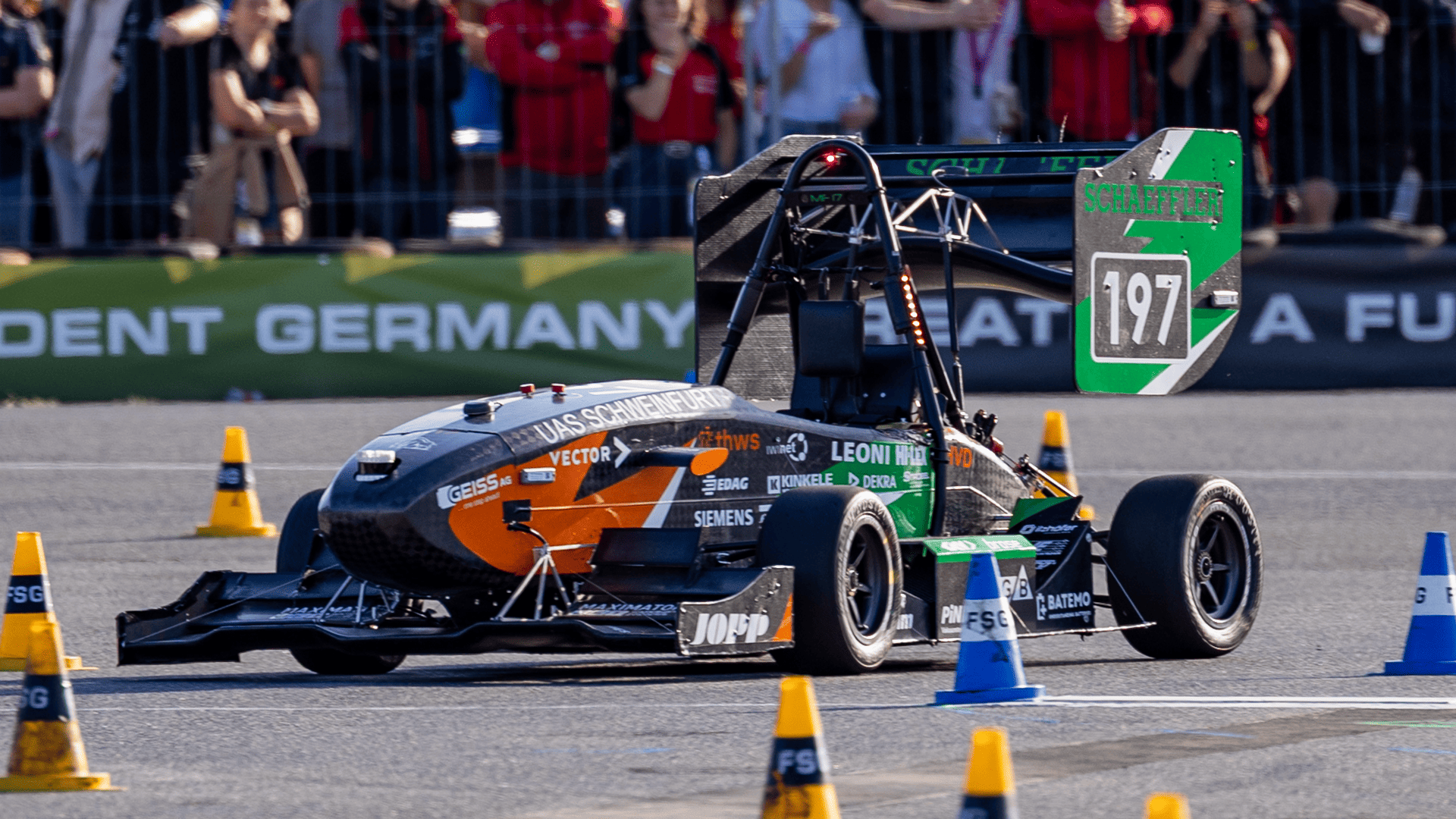}}
  \caption{Race car \textit{MF17} during autonomous driving on the competition Formula Student Germany. \textit{MF17} is a \SI{195}{\kilo\gram} electric rear wheel driven race car with a top speed of \SI{137}{\kilo\metre\per\hour}. Also visible are the blue and yellow cones indicating the track. © Formula Student Germany - Photo by lodholz}
  \label{fig:mf17}
  \vspace{-0.5 cm}
\end{figure}

%tell what this paper is about (Which problem)

To address these challenges, advanced control concepts such as \gls{mpc} are commonly used \cite{9790832}. Such controllers are based on prediction of future system behavior instead of just minimizing the current deviation from the target path. This enables optimal control even under dynamically changing conditions. For this reason, the quality of control largely depends on the accuracy of the underlying vehicle prediction model, as even small model deviations can lead to suboptimal behavior or instabilities \cite{Rawlings2020MPC}.

%How & what (How is it addressed) [Was hast du verändert? (anders als die anderen)]

This paper presents a novel vehicle prediction model for a Formula Student race car. It combines high prediction accuracy, robustness, real-time capability, and interpretability to ensure safety and meet the requirements of autonomous racing.
%
%Im Gegensatz zu einigen bestehenden Ansätzen kommt das vorgeschlagene Modell ohne Vorwissen über den Streckenverlauf aus und ist somit auch auf unbekannten Kursen einsetzbar. Es passt sich adaptiv an die aktuelle Fahrsituation an und ermöglicht dadurch eine robuste und präzise Modellierung des Fahrzeugverhaltens unter realen Wettbewerbsbedingungen.
%
%Main contribution (Explain what makes this work relevant - Was ist die Neuheit des Ansatzes?) [Wie hast du es verändert? (anders als die anderen)]
%
The main contribution of our approach lies in the strategic division of the model into three consecutive submodels: a nominal model, an offline model, and an online model. This modular structure combines physical prior knowledge with data-driven adaptation. It avoids the limitations of purely knowledge-based methods, which lack environmental adaptability, and of adaptive methods, which require excessive adjustment time.

The nominal model describes the physical properties and basic driving behavior of the vehicle based on a Kinematic Bicycle Model. This model provides suitable prediction accuracy at low velocities, but its accuracy decreases at higher velocities because of strong nonlinear forces that violate the model's geometric assumptions \cite{rajamani_vehicle_2012}. The offline model corrects these resulting residuals with information from previous runs and is realised using \gls{blr} \cite{10.1162/15324430152748236}. This allows the uncertainty of the prediction to be provided right from the start of the run and enables sufficient generalization across different environmental conditions. Finally, the online model, employing \gls{sgpr}, corrects remaining deviations between the model and actual vehicle behavior, caused e.g. by rain or different road surfaces. This allows the overall system to adaptively adjust the prediction model to the current environmental condition.
%
% Claims (Was hat die Neuheit Verbessert?)
%
By combining physical models with data-driven approaches, significantly improved prediction accuracy of the future vehicle state and information about the expected variance are achieved right from the start of the run. This knowledge significantly increases the robustness of subsequent control strategies.

Section \ref{sec:related_work} provides an overview of existing vehicle prediction models for path tracking \gls{mpc}s, followed by a detailed description of our approach in Section \ref{sec:implementation}. Finally, the performance of the approach is demonstrated in Section \ref{sec:evaluation} by directly comparing model accuracy and path tracking performance in an path tracking \gls{mpc}, with other commonly used prediction models. The \gls{mpc} is therefore introduced in Section \ref{sec:optimization_problem}. Lastly, the practical applicability of our approach is demonstrated using the race car \textit{MF17}. (see Fig. \ref{fig:mf17}).

%Da das Modell auf Daten aus früheren Fahrten zurückgreift, ist kein aufwändiges Expertenwissen über komplexe fahrzeugspezifische Parameter, wie detaillierte Reifenmodelle oder das Trägheitsmoment, erforderlich. Dies reduziert den Wartungsaufwand erheblich.

%Trotz der Einbeziehung umfangreicher Vorinformationen bleibt das Modell rechenzeiteffizient und erlaubt eine quantitative Einschätzung der Vorhersageunsicherheit über die Varianz. Gleichzeitig bleibt die Fähigkeit zur adaptiven Anpassung an aktuelle Fahrbedingungen, wie sie auch bei rein datengetriebenen Ansätzen üblich ist, vollständig erhalten.

%TODO kp ob man mehr auf das problem der echtzeitfähigkeit und performance eingehen sollte
%TODO nicht sicher ob man auch auf safty aware learning eingehen soll (lernweise muss sicher sein)
%TODO beschreibung Paperaufbau ????
%TODO evtl noch auf Laufzeit Anforderungen eingehen
%TODO evtl noch auf sicherheitsanforderungen eingehen 
%TODO teiweise schlechte argumentation. Lieber sagen keine gute Vorhersagegenauigkeit ab begin an und keine aussage über Varianz als offline daten nicht genutzt (lieber das mit offline daten als Lösung verkaufen für das problem)

	\section{Related Work}
	\label{sec:related_work}
%Put your paper into the scientific context

Due to the central role of the prediction model in path tracking \gls{mpc}s, a large amount of research on suitable vehicle prediction models has been conducted, both for road traffic and for autonomous racing. As this paper is situated within the context of Formula Student, we focus on the latter. For this reason, the presented prediction model is used in the area of path tracking \gls{mpc}s. It uses a reference path previously calculated by path planning and provides a target steering angle for the low-level control.

In the field of autonomous racing, the prediction model has to be reliable even at high speeds and with highly nonlinear driving behavior. Furthermore it has to offer high interpretability, as last-minute changes to the vehicle configuration may occur e.g. due to hardware defects. Complex, data-driven approaches such as neural networks are unsuitable, as small changes can lead to uninterpretable and potentially dangerous driving situations \cite{9536764, 9702227, 10611285}. Furthermore, it is usually not feasible to collect large, diverse data sets covering varying driving situations due to time constraints, which considerably limits the feasibility of data-based approaches \cite{8593882, alsunni2025llampcfastadaptivecontrol}. In addition, the small number of laps to be driven prevents the use of lap-intensive methods such as Iterative Learning Control \cite{8896988, brunke2020learningmodelpredictivecontrol}.

%What is the work previously done by others?

A common approach in \gls{mpc} for autonomous race cars is the use of physical vehicle models. While Tang \textit{et al.} \cite{9032103} propose a modified Kinematic Bicycle Model, Cataffo \textit{et al.} \cite{9945279} prefer a Dynamic Bicycle Model \cite{9341731, 9550038, 8167318}. Both models are computationally efficient, reducing runtime requirements, but at the expense of accuracy. This is primarily due to simplifying model assumptions and the fact that pure physical models do not allow for adaptation to current environmental conditions, such as rain. Our approach adapts the model to environmental conditions, improving prediction accuracy while correcting for errors from model assumptions. It also provides a quantitative measure of prediction uncertainty, improving the robustness of control.

The problem of adapting to current environmental conditions is usually solved in the literature by using \gls{sgpr} to learn and compensate for the residuals between the physical vehicle model and the actual values. Kabzan \textit{et al.} \cite{8754713} and Pinho \textit{et al.} \cite{wevj14070163} propose building a dictionary of residuals based on a scoring system in order to keep the runtime constant. \gls{sgpr} is then applied to this dictionary. This approach adapts to the current environmental condition and simultaneously learns the prediction variance, but requires a large amount of online data. As a result, high prediction accuracy is only available from the 2nd lap onwards, which can cause dangerous driving behavior in the 1st lap, leading, for example, to an emergency stop. 
%Our approach, on the other hand, provides high prediction accuracy and quantitative statements on the prediction variance right from the start.

Overall, compared to the current state of the art, the presented approach enables significantly higher prediction accuracy and a a quantitative measure of prediction uncertainty right from the start of the run. This is achieved through the efficient use of data from previous runs with sufficient abstraction, while maintaining real-time capability and key aspects for autonomous race cars, such as robustness and interpretability to ensure safety.

%Describe for every other paper, how your work differs

%TODO Summarize in which way your paper goes beyond the state of the art

% wegen den Iterative Learning Control sachen nochmal schauen ob man das überhaupt so erwähnt (ist halt eigentlich nur relativ schlecht wegargumentiert - das eine ist auch nur eine masterarbeit)
%Noch nicht verwendet:
% https://www.tandfonline.com/doi/full/10.1080/00423114.2022.2071300?scroll=top&needAccess=true (kein access - switcht iwie zwischen verschiedenen vehicle models via fuzzy logic)

% kp ob man die anderen modelle wie point mass model zitieren sollte (theoretisch auch noch nicht drinnen [LDM14])

	\section{Vehicle Prediction Model}
	\label{sec:implementation}
\begin{figure*}[htbp]
    \centering
    \vspace{0.2 cm}
    \def\svgwidth{1.0\textwidth}
    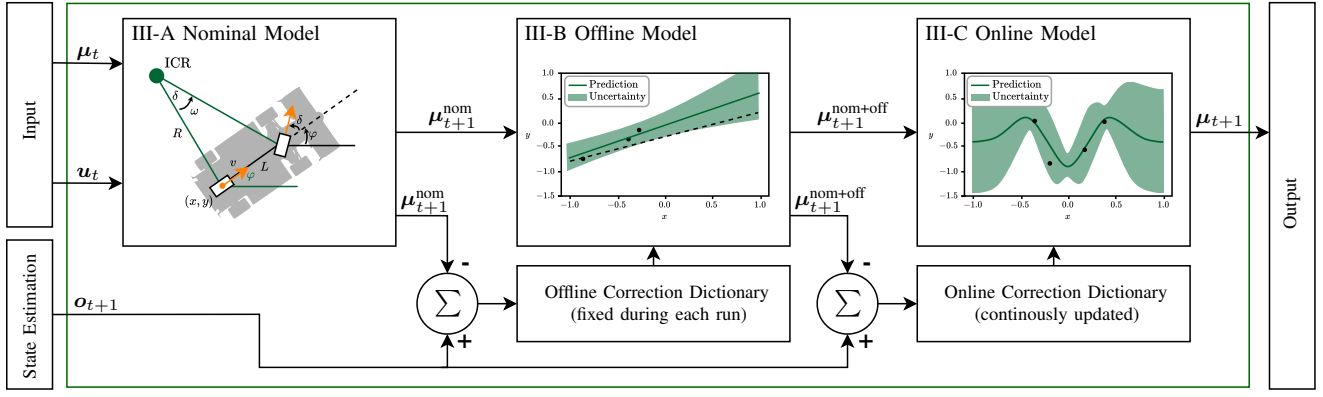
    \caption{Overview of our vehicle prediction model. It shows the three main components, Nominal Model (\ref{sec:nominal_vehicle_model}), Offline Model (\ref{sec:offline_model}), and Online Model (\ref{sec:online_model}), as well as the corresponding data dictionaries. In addition, the data flow can be seen.}
    \label{fig:prediction_model_overview}
    \vspace{-0.2 cm}
\end{figure*}

Our prediction model is illustrated in Fig. \ref{fig:prediction_model_overview}. The primary goal of the vehicle prediction model is to predict the future vehicle state $\vect{\mu}_{t+1}$ based on the current state $\vect{\mu}_t$ at time $t$. For a complete description of the vehicle state, this vector is defined as $\vect{\mu}_t = [x, y, \varphi, v_x, v_y, r, \delta, \Theta]^T_t$, where $x$ and $y$ represent the vehicle's position, and $\varphi$ the vehicle's orientation. $v_x$, $v_y$, and $r$ are the longitudinal, lateral, and rotational velocities. Furthermore a control command $\vect{u}_t = [\dot{\delta}, \dot{\Theta}]_t$ is given describing the change in the acceleration command $\Theta_t$ and in the steering angle command $\delta_t$.
To solve the prediction task we propose the use of three consecutive submodels: the nominal model, the offline model, and the online model. Due to its cascaded structure, where residuals are corrected in two consecutive steps, the overall system can be represented as the sum of these three submodels. Thus, the complete vehicle prediction model is obtained by
\begin{equation}
    \label{eq:full_prediction_model}
    \vect{\mu}_{t+1} = \vect{\mu}^\text{nom}_{t+1} + \vect{\mu}^\text{off}_{t+1} + \vect{\mu}^\text{on}_{t+1} \;\text{,}
\end{equation}
where $\vect{\mu}^\text{nom}_{t+1}$ denotes the prediction of the nominal model, $\vect{\mu}^\text{off}_{t+1}$ of the offline model, and $\vect{\mu}^\text{on}_{t+1}$ of the online model. 

Another key factor to achieve robust control is having knowledge about the uncertainty of the prediction. In our model, we approximate this by propagation of uncertainty using
\begin{equation}
    \label{eq:full_variance_model}
    \vect{\Sigma}_{t+1} = \vect{J}\vect{\Sigma}_t \vect{J}^T + \vect{\Sigma}^\text{off}_{t+1} + \vect{\Sigma}^\text{on}_{t+1} \;\text{.}
\end{equation}
Here, $\vect{\Sigma}_k$ denotes the covariance matrix of the current prediction step, $\vect{J}$ the jacobian matrix of the model, while $\vect{\Sigma}^\text{off}_{t+1}$ and $\vect{\Sigma}^\text{on}_{t+1}$ represent the covariance matrices of the models determined from the respective residuals \cite{Hewing_2020, Hewing_2018}.

\subsection{Nominal Model}
\label{sec:nominal_vehicle_model}

The nominal model forms the basis of our vehicle prediction model, enabling valid state predictions even without training data. It is based solely on physical properties. This paper proposes the use of a Kinematic Bicycle Model \cite{kabzan2019amzdriverlessautonomousracing}. Its key benefit is the small number of vehicle-specific parameters, primarily the distance from the front and rear axles to the center of gravity. In addition, the model is well defined at low speeds, which is not always guaranteed with models such as the Dynamic Bicycle Model \cite{rajamani_vehicle_2012, kabzan2019amzdriverlessautonomousracing}. The nominal model can be formulated as
\begin{equation}
    \label{eq:nominal_model}
\vect{\mu}^\text{nom}_{t+1} = 
\vect{\mu}_{t} + \begin{bmatrix}v_x \cos(\varphi) - v_y \sin(\varphi) \\ 
               v_x \sin(\varphi) + v_y \cos(\varphi) \\ 
               r \\ 
               \Theta \\ 
               (\dot{\delta} v_x + \delta \dot{v}_x) \frac{l_R}{l_R + l_F} \\ 
               (\dot{\delta} v_x + \delta \dot{v}_x) \frac{1}{l_R + l_F} \\
               \dot{\delta} \\
               \dot{\Theta}
\end{bmatrix}_t \cdot \Delta t \;\text{,}
\end{equation}
where $l_R$ and $l_F$ denote the distances from the vehicle's center of gravity to the rear and front axles.

\subsection{Offline Model Correction}
\label{sec:offline_model}
\glsreset{blr}

Since a physical model, like the nominal model, is limited in its accuracy due to its model assumptions, the inclusion of data from previous runs offers great potential for improving the prediction quality. This is particularly important at higher speeds, as the Kinematic Bicycle Model assumes negligible tire slip and therefore loses accuracy within this operating range \cite{rajamani_vehicle_2012}. Furthermore, the data gathered from previous runs can be used to learn the uncertainties of the nominal model to assess its quality. 

The mapping matrix $\vect{B}$ allows to extract the velocities from the state and is defined by $\vect{B} = [\vect{0}_{3\times3}, \vect{I}_{3\times3}, \vect{0}_{3\times2}]$. This is necessary since the correction models only learn to output the velocity residuals. These residuals sufficiently compensate for all model errors through the integrated prediction of position and orientation. To simplify the learning problem, the offline model only learns the residuals $\vect{\epsilon}_t^\text{off} \in \mathbb{R}^3$ between the predictions of the nominal model and the actual measured states. These residuals are defined by
\begin{equation}
    \vect{\epsilon}_t^\text{off} = \vect{B} \left(\vect{o}_{t+1}-\vect{\mu}_{t+1}^\text{nom} \right) \;\text{,}
\end{equation}
where $\vect{o}_{t+1}$ denotes the current measurement of the state variables $\vect{\mu}_t$. Learning a function of these residuals in relation to the system state $\vect{s}_t = [\vect{\mu}, \vect{u}]^T_t$ requires a machine learning approach with a high transferability to unknown situations. For this purpose, we propose the use of a \gls{blr} \cite{10.1162/15324430152748236}, which takes velocities, steering angle and control command as features. This method is capable of generalizing between different driving situations and captures its uncertainties. Furthermore, it requires minimal data, provides interpretable results, and can therefore be safely used in vehicle-related applications. The \gls{blr} is applied on a previously gathered dictionary, which consists of $N$ tuples $\langle\vect{s}_t, \vect{\epsilon}_t^\text{off}\rangle_n$ from previous runs and is only updated once prior to the run. As a result, the offline model can be defined as
\begin{equation}
    \label{eq:offline_model}
    \begin{split}
        \vect{\mu}_{t+1}^\text{off} &= \vect{B}^T \mathrm{clamp}\! \left(\mathrm{BLR}_\mu \left(\vect{s}_t | \big\{ \langle\vect{s}_t, \vect{\epsilon}_t^\text{off}\rangle_n \big\}^N_{n=1} \right), \vect{\tau}\right) \text{,}\\
        \vect{\Sigma}_{t+1}^\text{off} &= \vect{B}^T \mathrm{BLR}_\Sigma \left(\vect{s}_t|\big\{ \langle\vect{s}_t, \vect{\epsilon}_t^\text{off}\rangle_n \big\}^N_{n=1} \right) \vect{B} \;\text{.}
    \end{split}
\end{equation}
Since strong nonlinear effects occur at high speeds, the offline model can only correct a limited range of the state space with a linear model reliably. Therefore, it is necessary to introduce a trust region. For this purpose the prediction of the mean value is adjusted with a clamping function
\begin{equation}
    \mathrm{clamp}(\vect{x}, \vect{\tau}) = \min(\max(\vect{x}, -\vect{\tau}), \vect{\tau}) \;\text{,}
\end{equation}
where $\vect{\tau}$ denotes a threshold for the corresponding value. Assuming the residuals are normally distributed, this value can be determined through the 99th percentile of the residuals.

\subsection{Online Model Correction}
\label{sec:online_model}

To compensate for the remaining model errors caused by the general nature of the nominal and offline models and their inability to adapt to the current environmental conditions, an online model is introduced for real-time compensation. This is achieved by using a \gls{sgpr} with an \gls{rbf} kernel. This approach enables non-parametric regression, which allows a precise modeling of any shape of the remaining residuals $\vect{\epsilon}^\text{on}_t \in \mathbb{R}^3$ that are defined by
\begin{equation}
    \vect{\epsilon}^\text{on}_t = \vect{B} \big(\vect{o}_{t+1}-\underbrace{\big(\vect{\mu}_{t+1}^\text{nom} + \vect{\mu}_{t+1}^\text{off} \big)}_{\vect{\mu}_{t+1}^\text{nom+off}} \big) \;\text{,}
\end{equation}
while simultaneously meeting the real-time requirements of the system. Additionally, no abstraction capabilities are necessary since the online model is only used during the ongoing run. In this way, the online model can be defined by
\begin{equation}
    \label{eq:online_model}
    \begin{split}
        \vect{\mu}_{t+1}^\text{on} &= \vect{B}^T \mathrm{SGPR}_\mu \left(\vect{s}_t|\big\{ \langle\vect{s}_t, \vect{\epsilon}^\text{on}_t \rangle_m \big\}^M_{m=1}, \vect{P}_t \right) \text{,}\\
        \vect{\Sigma}_{t+1}^\text{on} &= \vect{B}^T \mathrm{SGPR}_\Sigma \left(\vect{s}_t|\big\{ \langle\vect{s}_t, \vect{\epsilon}^\text{on}_t \rangle_m \big\}^M_{m=1}, \vect{P}_t \right) \vect{B} \;\text{.}
    \end{split}
\end{equation}
For this purpose, the inducing points $\vect{P}_t$ should be set as close as possible to the expected states e.g. based on a previous \gls{mpc} prediction \cite{8754713}. A separate \gls{sgpr} is then applied for each target variable, using velocity, steering angle, and throttle command as input features.

Limiting the runtime of the model is a crucial aspect for the real-time capability of the approach. Because of this a dictionary of size $M$ is used to create a manageable and representative set of residuals from the continuous online data stream. This dictionary is used to store and update relevant data points in a structured manner. To decide whether a new data point should be added to the dictionary, a score is calculated that evaluates the relevance of the current measured value \cite{8754713}. This results in the insertion condition
\begin{equation}
    \lambda_{\text{max}} \left(\vect{\Sigma}_{t+1}^\text{on} \right) > \eta_\text{insertion} \;\text{,}
\end{equation}
with $\lambda_{\text{max}} \left(\vect{\Sigma}_{t+1}^\text{on} \right)$ describing the largest eigenvalue of the matrix $\vect{\Sigma}_{t+1}^\text{on}$, while $\eta_\text{insertion}$ represents an empirically chosen threshold. Once the maximum number of entries $M$ in the dictionary has been reached, the data point $\hat{m}$, whose associated score has the smallest value is removed. This point can be found by 
\begin{equation}
    \label{eq:replace_calculation}
    \hat{m} = \arg\min_m \left( \exp\left( - \frac{\left(t - t_m \right)^2}{2 \eta_\text{time}} \right) \cdot \lambda_{\text{max}} \! \left(\vect{\Sigma}_m^\text{off} \right) \right) \;\text{.}
\end{equation}
Here, we incorporate the age of the residuals to ensure that older data points are removed preferentially. In (\ref{eq:replace_calculation}), $t$ denotes the current timestamp and $t_m$ the time at which the respective data point $m$ was recorded, while $\xi_\text{time}$ represents a parametric weighting factor, controlling the influence of age. To ensure the safety of the learning process and prevent any endangerment of the vehicle due to unrealistically high residuals caused by measurement errors, an additional outlier condition, $||\vect{\epsilon}^\text{on}_t||^2 \leq \xi_\text{outlier}$ is defined. Here a empirically chosen threshold $\xi_\text{outlier}$ can be used, to ensure that only valid data points are included in the model.

	\section{Optimization Problem}
	\label{sec:optimization_problem}

Our proposed vehicle prediction model can then be used in an \gls{mpc} for example. For this purpose, our prediction model presented in Section \ref{sec:implementation} can be combined using (\ref{eq:full_prediction_model}) and (\ref{eq:full_variance_model}), where each submodel is dependent on $\vect{u}_t$ through the use of $\dot{\delta}$, $\dot{\Theta}$ or $\vect{s}$. The main objective of the controller is to minimize the distance between the predicted path and the reference path $\vect{\Gamma}$, where the error at each prediction step $k$ is represented by
\begin{equation}
    \begin{split}
        \begin{bmatrix}e_l \\ e_c\end{bmatrix}_k =
        \vect{R}(\Gamma_{\varphi, k}) \begin{bmatrix} \mu_{x} - \Gamma_{x} \\ \mu_{y} - \Gamma_{y}\end{bmatrix}_k \;\text{.}
    \end{split}
\end{equation}
Therefore, one reference point $\vect{\Gamma}_k = [x, y, \varphi, w]_k$ consists of a reference position $(x, y)$, the path orientation $\varphi$ indicating the direction of travel, and the track width $w$. Furthermore $\vect{R}(\Gamma_{\varphi, k})$ represents a clockwise rotation matrix. The positions of the reference points are selected based on the target speed, previously obtained from a velocity planning module \cite{Heilmeier02102020}. They represent the nearest points on the track to the respective vehicle positions. In general, the controller receives the target path and the current vehicle state, as well as the target velocity profile for the next few meters, as inputs. The \gls{mpc} then minimizes the cost function by adjusting the sequence of control commands $\vect{U} = [\vect{u}_1, \ldots, \vect{u}_K]$, resulting in an optimal control command sequence. Based on this, the controller solves the following optimization problem:
\begin{equation}
    \begin{split}
        \min_{\vect{U}} &\quad \frac{1}{K} \sum_{k=1}^K \left(\eta_c e_{k, c}^2 + \eta_l e_{k, l}^2 + \eta_s \dot{\delta}_k^2 \right) \\
        \text{s.t.} &\quad \vect{\mu}_0 = \vect{o}_{t} \\
                    &\quad \vect{\mu}_{k+1} \overset{\text{(\ref{eq:full_prediction_model})}}{=} \vect{\mu}^\text{nom}_{k+1} + \vect{\mu}^\text{off}_{k+1} + \vect{\mu}^\text{on}_{k+1} \\
                    &\quad \vect{\Sigma}_{k+1} \overset{\text{(\ref{eq:full_variance_model})}}{=} \vect{J}\vect{\Sigma}_k \vect{J}^T + \vect{\Sigma}^\text{off}_{k+1} + \vect{\Sigma}^\text{on}_{k+1}  \\
                    &\quad \delta_\text{min} \leq \mu_{k,\delta} \leq \delta_\text{max}\\
                    &\quad e_{k, c}^2 \leq \left(\Gamma_{k, w} - \sqrt{\chi^2_2(\eta_p) \lambda_{\text{max}} (\vect{\Sigma}_{k})} \right)^2 \;\text{.}\\
    \end{split}
\end{equation}
We solved this using \gls{sqp} \cite{Boggs_Tolle_1995}. Here, the weighting factors of the cost function are represented by the parameters $\eta_c$, $\eta_l$, and $\eta_s$, while the maximum allowed probability of violating the track limits is defined by the parameter $\eta_p$ \cite{Hewing_2018}. Furthermore a steering constraint integrates the physical limits of the steering system into the optimization problem with $\delta_\text{min}$ and $\delta_\text{max}$ indicating the maximum steering angle. Overall, all error terms $\vect{e_{k}}$ as well as the vehicle states $\vect{\mu}_{k}$ are dependent on $\vect{U}$ through the use of our proposed submodels (\ref{eq:nominal_model}), (\ref{eq:offline_model}) and (\ref{eq:online_model}).

	\section{Experiments}
	\label{sec:evaluation}
%\glsreset{mpc}

To demonstrate the actual performance of our vehicle prediction model under realistic conditions, we compare its path tracking performance with common approaches. These include a pure physical model \cite{9945279} and a \gls{gpr} corrected model \cite{8754713}, as described in Section \ref{sec:related_work}. In addition, the performance of the individual submodels are analyzed and the applicability of our model to a real Formula Student race car is investigated. For these experiments, we use the simple path tracking MPC of Section \ref{sec:optimization_problem}, where only the first steering angle command is applied. For this purpose, we set the parameters of the controller to $\eta_c=100$, $\eta_l=30$, $\eta_s=10$, and $p=0.95$. Furthermore, a sampling time of $\Delta t=$ \SI{0.1}{\second} with $K=10$ prediction steps was chosen for the experiments. The acceleration is controlled by a \gls{pid} controller to achieve the desired velocity. 

In particular, prediction accuracy and runtime are used as evaluation criteria. The performance of the model can be represented by the prediction error calculated by 
\begin{equation}
  e_\text{pred} = |\vect{B} (\vect{o}_{t+1} - \vect{\mu}_{t+1})| \;\text{,}
\end{equation}
and by the path error describing the euclidean distance
\begin{equation}
  e_\text{path} = \sqrt{({o}_{t, x} - {\Gamma}_{t,x})^2 + ({o}_{t, y} - {\Gamma}_{t, y})^2} \;\text{,}
\end{equation}
where ${\Gamma}_{t,x}$ and ${\Gamma}_{t, y}$ describe the nearest point on the reference path to the current vehicle position. Both metrics provide information about the accuracy of the respective prediction model.

\subsection{Evaluation of the Submodels}
%\glsreset{sgpr}
%\glsreset{blr}

First, the individual submodels are analyzed separately to access their performance. Various model variants are tested using real-world vehicle data. Therefore, the training and test data originates from different weather conditions, racetracks, and asphalt types. Initially, a pure offline model consisting of the nominal model with offline correction is examined. The corresponding prediction errors $e_\text{pred}$ are shown in Table \ref{tab:offline_model_comparison}.
\begin{table}[htbp]
\caption{Offline Model Prediction Error Comparison}
\begin{center}
    \begin{tabular}{|c|c|c|c|}
        
    \hline
    %\textbf{Model} & \textbf{\textit{Median}} & \textbf{\textit{Max}}& \textbf{\textit{$\bm{Q_{0.75}}$}} \\
    \textbf{Model} & \textbf{\textit{Median}} & \textbf{\textit{Max}}& \textbf{\textit{$\bm{Q_{0.75}}$}} \\
    \hline
    Kinematic Bicycle Model              & 0.445    & 2.439 & 0.117 \\ % (0 \%)
    \hline
    \textbf{Kinematic Bicycle Model + BLR} & 0.186  & 4.134 & 0.084 \\ % (58 \%)
    \hline
    Kinematic Bicycle Model + SGPR         & 0.550 & 5.282 & 0.328 \\ % (-23 \%)
    \hline
    Dynamic Bicycle Model                & 0.244  & 2.437 & 0.148 \\ % (45 \%)
    \hline
    Dynamic Bicycle Model + BLR            & 0.149  & 3.875 & 0.101 \\ % (66 \%)
    \hline
    Dynamic Bicycle Model + SGPR           & 0.420   & 6.139 & 0.253 \\ % (5 \%)
    \hline

    \end{tabular}
    \label{tab:offline_model_comparison}
\end{center}
\end{table}
It can be seen that the Dynamic Bicycle Model achieves the best performance as a nominal model in combination with a \gls{blr} as offline model. Close behind is the Kinematic Bicycle Model with \gls{blr} offline model, showing only a slight difference in accuracy. Due to its better model definition at low speeds, the Kinematic Bicycle Model is a particularly suitable option for our usecase. It is also noticeable that the Kinematic Bicycle Model with \gls{blr} correction provides more accurate predictions than the pure Dynamic Bicycle Model. In contrast, the \gls{sgpr} models perform comparatively poorly. This is due to strong overfitting to the training data, as a large dataset is needed to learn a general representation under all conditions. Consequently, the model's ability to abstract is limited with small datasets, which are common in Formula Student applications. This significantly worsens the accuracy of predictions for conditions that differ from the training data.

\begin{figure}[htbp]
  \centering
  %\vspace{0.3 cm}
  \resizebox{\columnwidth}{!}{\input{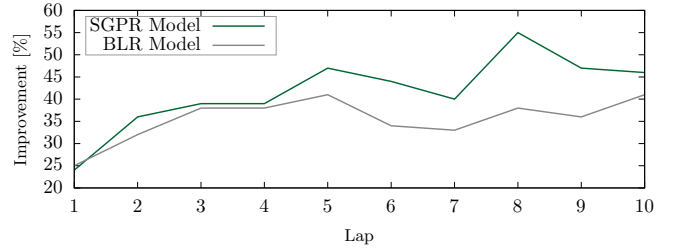}}
  \caption{Comparison of different online models in terms of improvement in prediction error $e_\text{pred}$ compared to a pure Kinematic Bicycle Model. A \gls{sgpr} (green) and a \gls{blr} (gray) were compared as online model candidates over 10 laps. As nominal and offline model a Kinematic Bicycle Model with a \gls{blr} was used.}
  \label{fig:online_model_comparison}
\end{figure}
Fig. \ref{fig:online_model_comparison}, shows the effects of different online models. Here, an opposite phenomenon can be observed. While the \gls{blr} reaches its maximum learning potential by the 3rd lap and achieves no further improvements, the \gls{sgpr} model continues to improve. This can be attributed to the fact that the \gls{blr} reaches a natural limit due to its linear representation of residuals. In contrast, the \gls{sgpr} can continue to overfit the training data. In this case, this is actually an advantageous effect since the model is only used for the current run and therefore only needs to be precise under current conditions.

\subsection{Comparison of Path Tracking Performance}

\begin{figure}[htbp]
  \centering
  \resizebox{\columnwidth}{!}{\input{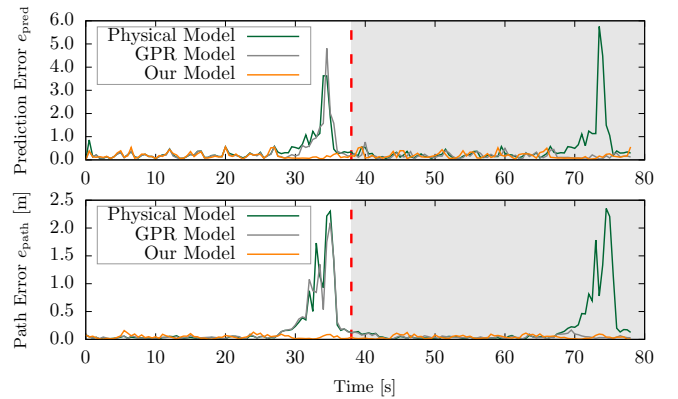}}
  \caption{Comparison of different vehicle prediction models in terms of prediction error $e_\text{pred}$ and path tracking error $e_\text{path}$. We compare to the physical model \cite{9945279} and the \gls{gpr} corrected model \cite{8754713}. Two laps were driven, with the lap change marked by the red dashed line.}
  \label{fig:model_path_error}
\end{figure}

In order to better access the accuracy of our full vehicle prediction model (see Fig. \ref{fig:prediction_model_overview}) under realistic operating conditions, the model was additionally integrated in a path tracking \gls{mpc} for evaluation. For this purpose, 2 laps were driven in a simulation, deliberately creating the most challenging driving conditions possible, by having a low tire-road friction coefficient set. Fig. \ref{fig:model_path_error} shows the prediction error $e_\text{pred}$ and the resulting path error $e_\text{path}$ of our model, which is compared to the common approaches of a pure physical model \cite{9945279} and the \gls{gpr} corrected model \cite{8754713} from Section \ref{sec:related_work}.

\begin{figure}[htbp]
  \centering
  \vspace{0.3 cm}
  \resizebox{\columnwidth}{!}{\input{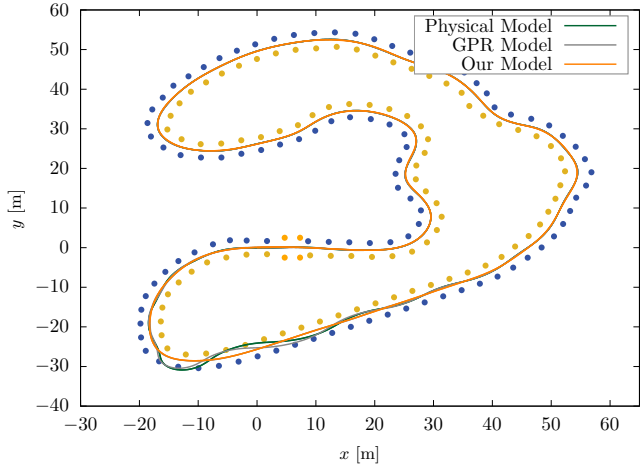}}
  \caption{Comparison of different vehicle prediction models in relation to the driven path, with the points representing the track boundaries. We compare to the physical model \cite{9945279} and the \gls{gpr} corrected model \cite{8754713}. Two laps were driven.}
  \label{fig:trajectory}
  \vspace{-0.5 cm}
\end{figure}
In addition, Fig. \ref{fig:trajectory} shows a corresponding visualization of the actual path driven. Overall, the expected strong correlation between prediction error and path error was observed. For the nominal model, it can be seen that the vehicle begins to oscillate when accelerating out of the curve onto the long straight at (-10, 40)[\si{\meter}]. It is also noticeable that this oscillation causes the system to enter highly nonlinear and unknown areas of the vehicle prediction model, further increasing prediction errors. The \gls{gpr} model exhibits similar behavior in the 1st lap. Due to the learning ability of the \gls{gpr} model, however, a significant improvement can be seen from the 2nd lap onwards, resulting in a more stable driving style. In contrast to these two models, ours shows a very stable driving style from the beginning and delivers even in the 2nd lap slightly better results than the other models. Overall, a reduction in prediction error of \SI{57}{\percent} compared to the physical model and \SI{32}{\percent} compared to the \gls{gpr} model can be measured.

\begin{figure}[htbp]
  \centering
  \resizebox{\columnwidth}{!}{\input{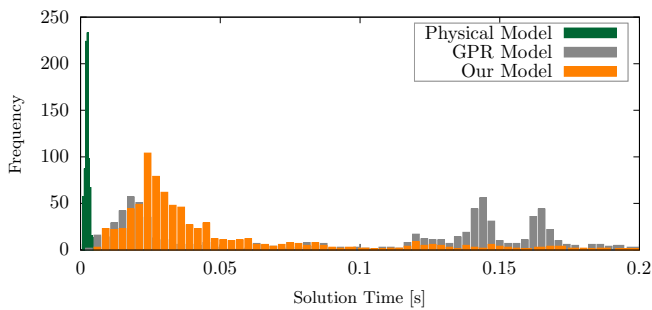}}
  \caption{Comparison of different vehicle prediction models based on the \gls{mpc} runtime. The physical model \cite{9945279} and the \gls{gpr} corrected model \cite{8754713} was used for comparison. The figure shows the accumulated times on a two-lap run.}
  \label{fig:runtime}
\end{figure}

Another crucial factor is the runtime of the vehicle prediction model to ensure the real-time capability of the system. Fig. \ref{fig:runtime} shows the corresponding runtime of the \gls{mpc} for the various models.
It can be seen that the physical model takes the least time to solve. Our model is also faster than the \gls{gpr} corrected model. This can be attributed to the absence of oscillations, which lead to highly nonlinear behavior that makes the optimization problem more difficult. Since the \gls{mpc} uses the result of the last optimization as the new starting condition, the optimizer requires fewer adjustments and thus converges faster, if the model is accurate.

\subsection{Experiments on the Real Vehicle}

\begin{figure}[htbp]
  \centering
  \resizebox{\columnwidth}{!}{\input{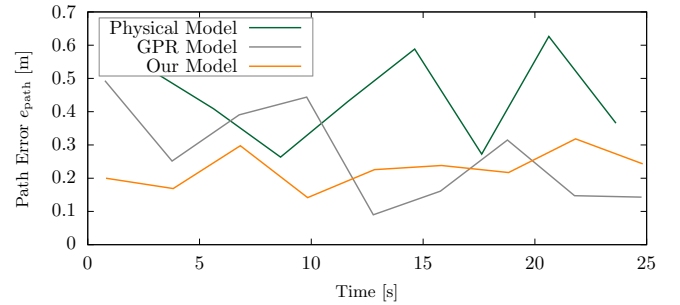}}
  \caption{Comparison of different prediction models based on the path tracking behavior on the real Formula Student race car \textit{MF17}. We compare to the physical model \cite{9945279} and the \gls{gpr} corrected model \cite{8754713}}
  \label{fig:real_prediction_model_comparison}
\end{figure}

Finally, it can be shown that good results can also be achieved on a real race car. For this purpose, one lap was driven with the \textit{MF17} (Fig. \ref{fig:mf17}). Therefore, we compare our model to the physical model \cite{9945279} and the \gls{gpr} corrected model \cite{8754713}. Fig. \ref{fig:real_prediction_model_comparison} shows the corresponding evaluation of the path tracking error.
%
%\begin{figure}[htbp]
%  \centerline{\includegraphics[width=0.5\textwidth]{plots/real_prediction_model_comparison.pdf}}
%  \caption{Comparison of different prediction models based on the path tracking behavior on the real Formula Student race car \textit{MF17}. The physical model approach \cite{9945279} and the \gls{gpr} model approach \cite{8754713} was used for this purpose.}
%  \label{fig:real_prediction_model_comparison}
%\end{figure}
From the very beginning, our model demonstrates significantly lower errors compared to the other approaches, thus offering sufficient path tracking performance. The \gls{gpr} corrected model only reaches a similar performance level after some time, while the physical model consistently delivers poorer results. Overall, this confirms the results obtained in the simulation. Furthermore, a valid path tracking performance can be seen, thus validating the applicability of our approach on a real race car.

	\section{Conclusion}
	\glsresetall
% Again describe the approach presented in this paper
%
This paper presented a novel vehicle prediction model for path tracking with \gls{mpc} in the context of Formula Student Driverless. The novelty of the model lies in its subdevision into three consecutive submodels consisting of a nominal, an offline, and an online model which enables efficient integration of all available data. Other approaches only use a subset of these components.
%
% Again mention the advantages and what is noval compared to previous approaches
%
The combination of such models ensures high prediction quality and quantitative assessment of the model uncertainty right from the start of the run, while maintaining the system's adaptability to the current situation and real-time capability.
%
% Mention the implementation and the successful outcome of the experiments

An improvement of up to \SI{57}{\percent} in both prediction quality and path tracking accuracy was successfully demonstrated compared to common approaches, such as purely physical models or \gls{gpr} correction-based models. Our model's performance improvement compared to common approaches was particularly evident in the 1st lap. The \gls{gpr} corrected model required one lap to achieve a comparable level of performance, which can lead in the worst case to dangerous behavior or even crashes on the 1st lap. We further demonstrated the applicability of our approach using a real Formula Student race car, where it showed the best results.

% Potentially discuss options for future work

We see significant potential for future work in applying the presented approach to other areas, such as state estimation or other \gls{mpc} variants. Moreover, our approach could be applied beyond autonomous racing, for instance, in regular road traffic.
	
	%%%%%%%%%%%%%%%%%%%%%%%%%%%%%%%%%%%%%%%%%%%%%%%%%%%%%%%%%%%%%%%%%%
	%\section*{ACKNOWLEDGMENTS}
	%This work was done in cooperation with Mainfranken Racing e.V. %. Support was provided in evaluation and driving dynamics expertise.
	
	%%%%%%%%%%%%%%%%%%%%%%%%%%%%%%%%%%%%%%%%%%%%%%%%%%%%%%%%%%%%%%%%%%
	%\addtolength{\textheight}{-12cm}
	%\vspace{10mm}
	\bibliographystyle{IEEEtran}
	% Your .bib file here
	\bibliography{references} 

@INPROCEEDINGS{9945279,
  author={Cataffo, Vittorio and Silano, Giuseppe and Iannelli, Luigi and Puig, Vicenç and Glielmo, Luigi},
  booktitle={2022 IEEE International Conference on Systems, Man, and Cybernetics (SMC)}, 
  title={A Nonlinear Model Predictive Control Strategy for Autonomous Racing of Scale Vehicles}, 
  year={2022},
  volume={},
  number={},
  pages={100-105},
}

@INPROCEEDINGS{9341731,
  author={Vázquez, José L. and Brühlmeier, Marius and Liniger, Alexander and Rupenyan, Alisa and Lygeros, John},
  booktitle={2020 IEEE/RSJ International Conference on Intelligent Robots and Systems (IROS)}, 
  title={Optimization-Based Hierarchical Motion Planning for Autonomous Racing}, 
  year={2020},
  volume={},
  number={},
  pages={2397-2403},
}

@INPROCEEDINGS{9550038,
  author={Liu, Yanbo and Sun, Weiqi and Liu, Shiji and Xu, Wenchao and Xiong, Xin and Hao, Li and Hu, Fangjie and Qu, Ling},
  booktitle={2021 40th Chinese Control Conference (CCC)}, 
  title={Dynamic Path Planning for Formula Automous Racing Cars}, 
  year={2021},
  volume={},
  number={},
  pages={6087-6093},
}

@INPROCEEDINGS{9702227,
  author={Kebbati, Yassine and Ait-Oufroukh, Naima and Vigneron, Vincent and Ichalal, Dalil},
  booktitle={2021 IEEE International Conference on Networking, Sensing and Control (ICNSC)}, 
  title={Neural Network and ANFIS based auto-adaptive MPC for path tracking in autonomous vehicles}, 
  year={2021},
  volume={1},
  number={},
  pages={1-6},
}

@INPROCEEDINGS{10611285,
  author={Gomes, David R. and Botto, Miguel Ayala and Lima, Pedro U.},
  booktitle={2024 IEEE International Conference on Robotics and Automation (ICRA)}, 
  title={Learning-based Model Predictive Control for an Autonomous Formula Student Racing Car}, 
  year={2024},
  volume={},
  number={},
  pages={12556-12562},
}

@INPROCEEDINGS{8593882,
  author={McKinnon, Christopher D. and Schoellig, Angela P.},
  booktitle={2018 IEEE/RSJ International Conference on Intelligent Robots and Systems (IROS)}, 
  title={Experience-Based Model Selection to Enable Long-Term, Safe Control for Repetitive Tasks Under Changing Conditions}, 
  year={2018},
  volume={},
  number={},
  pages={2977-2984},
}

@INPROCEEDINGS{Hewing_2018,
  author={Hewing, Lukas and Liniger, Alexander and Zeilinger, Melanie N.},
  booktitle={2018 European Control Conference (ECC)}, 
  title={Cautious NMPC with Gaussian Process Dynamics for Autonomous Miniature Race Cars}, 
  year={2018},
  volume={},
  number={},
  pages={1341-1348},
}

@ARTICLE{8167318,
  author={Liniger, Alexander and Lygeros, John},
  journal={IEEE Transactions on Control Systems Technology}, 
  title={Real-Time Control for Autonomous Racing Based on Viability Theory}, 
  year={2019},
  volume={27},
  number={2},
  pages={464-478},
}

@ARTICLE{9032103,
  author={Tang, Luqi and Yan, Fuwu and Zou, Bin and Wang, Kewei and Lv, Chen},
  journal={IEEE Access}, 
  title={An Improved Kinematic Model Predictive Control for High-Speed Path Tracking of Autonomous Vehicles}, 
  year={2020},
  volume={8},
  number={},
  pages={51400-51413},
}

@ARTICLE{8754713,
  author={Kabzan, Juraj and Hewing, Lukas and Liniger, Alexander and Zeilinger, Melanie N.},
  journal={IEEE Robotics and Automation Letters}, 
  title={Learning-Based Model Predictive Control for Autonomous Racing}, 
  year={2019},
  volume={4},
  number={4},
  pages={3363-3370},
}

@ARTICLE{wevj14070163,
  author = {Pinho, João and Costa, Gabriel and Lima, Pedro U. and Ayala Botto, Miguel},
  journal = {World Electric Vehicle Journal},
  title = {Learning-Based Model Predictive Control for Autonomous Racing},
  year = {2023},
  volume = {14},
  number = {7},
  pages={},
}

@ARTICLE{9536764,
  author={Rokonuzzaman, Mohammad and Mohajer, Navid and Nahavandi, Saeid and Mohamed, Shady},
  journal={IEEE Access}, 
  title={Model Predictive Control With Learned Vehicle Dynamics for Autonomous Vehicle Path Tracking}, 
  year={2021},
  volume={9},
  number={},
  pages={128233-128249},
}

@ARTICLE{8896988,
  author={Rosolia, Ugo and Borrelli, Francesco},
  journal={IEEE Transactions on Control Systems Technology}, 
  title={Learning How to Autonomously Race a Car: A Predictive Control Approach}, 
  year={2020},
  volume={28},
  number={6},
  pages={2713-2719},
}

@ARTICLE{9790832,
  author={Betz, Johannes and Zheng, Hongrui and Liniger, Alexander and Rosolia, Ugo and Karle, Phillip and Behl, Madhur and Krovi, Venkat and Mangharam, Rahul},
  journal={IEEE Open Journal of Intelligent Transportation Systems}, 
  title={Autonomous Vehicles on the Edge: A Survey on Autonomous Vehicle Racing}, 
  year={2022},
  volume={3},
  number={},
  pages={458-488},
}

@ARTICLE{Hewing_2020,
  author={Hewing, Lukas and Kabzan, Juraj and Zeilinger, Melanie N.},
  journal={IEEE Transactions on Control Systems Technology}, 
  title={Cautious Model Predictive Control Using Gaussian Process Regression}, 
  year={2020},
  volume={28},
  number={6},
  pages = {2736-2743},
}

@ARTICLE{10.1162/15324430152748236,
  author = {Tipping, Michael E.},
  journal = {The Journal of Machine Learning Research},
	title = {Sparse bayesian learning and the relevance vector machine},
	year={2001},
  volume = {1},
  number={},
  pages = {211-244},
}

@ARTICLE{Boggs_Tolle_1995,
  author = {Boggs, Paul T. and Tolle, Jon W.},
  journal = {Acta Numerica},
	title = {Sequential Quadratic Programming},
	year={1995},
  volume = {4},
  number={},
  pages = {1-51},
}

@ARTICLE{kabzan2019amzdriverlessautonomousracing,
  author = {Kabzan, Juraj and Valls, Miguel I. and Reijgwart, Victor J. F. and Hendrikx, Hubertus F. C. and Ehmke, Claas and Prajapat, Manish and Bühler, Andreas and Gosala, Nikhil and Gupta, Mehak and Sivanesan, Ramya and Dhall, Ankit and Chisari, Eugenio and Karnchanachari, Napat and Brits, Sonja and Dangel, Manuel and Sa, Inkyu and Dubé, Renaud and Gawel, Abel and Pfeiffer, Mark and Liniger, Alexander and Lygeros, John and Siegwart, Roland},
  journal = {Journal of Field Robotics},
	title = {AMZ Driverless: The full autonomous racing system},
	year={2020},
  volume = {37},
  number={7},
  pages = {1267-1294},
}

@ARTICLE{Heilmeier02102020,
  author = {Heilmeier, Alexander and Wischnewski, Alexander and Hermansdorfer, Leonhard and Betz, Johannes and Lienkamp, Markus and Lohmann, Boris},
  journal = {Vehicle System Dynamics},
	title = {Minimum curvature trajectory planning and control for an autonomous race car},
	year={2020},
  volume = {58},
  number={10},
  pages = {1497-1527},
}

@misc{alsunni2025llampcfastadaptivecontrol,
  title={LLA-MPC: Fast Adaptive Control for Autonomous Racing}, 
  author={Maitham F. AL-Sunni and Hassan Almubarak and Katherine Horng and John M. Dolan},
  year={2025},
  eprint={2505.19512},
  archivePrefix={arXiv},
  primaryClass={cs.RO},
  note={arXiv:2505.19512 [cs.RO]},
  doi={10.48550/arXiv.2505.19512}, 
}

@misc{brunke2020learningmodelpredictivecontrol,
  title={Learning Model Predictive Control for Competitive Autonomous Racing}, 
  author={Lukas Brunke},
  year={2020},
  eprint={2005.00826},
  archivePrefix={arXiv},
  primaryClass={cs.LG},
  note={arXiv:2005.00826 [cs.LG]},
  doi={10.48550/arXiv.2005.00826},
}

@online{fsg_rules,
  author    = {Formula Student Germany},
  shortauthor = {FSG},
  title     = {Formula Student Rules 2026},
  year      = {2025},
  url       = {https://www.formulastudent.de/fsg/rules/},
  urldate   = {2025-10-17} 
}

@book{Rawlings2020MPC,
  title     = {Model Predictive Control: Theory and Design},
  author    = {James B. Rawlings and David Q. Mayne and Moritz M. Diehl},
  year      = {2017},
  edition   = {2nd},
  publisher = {Nob Hill Publishing},
}

@book{rajamani_vehicle_2012,
  title = {Vehicle Dynamics and Control},
  author = {Rajamani, Rajesh},
  year = {2012},
  edition   = {2nd},
  publisher = {Springer US},
}
	
\end{document}